# X-Ray cardiac angiographic vessel segmentation based on pixel classification using machine learning and region growing

**E O Rodrigues[1], L O Rodrigues[2], J J Lima[1], D Casanova[1], F Favarim[1], E R Dosciatti[1], V Pegorini[1], L S N Oliveira[3] and F F C Morais[4]**

[1] Department of Academic Informatics (DAINF), Universidade Tecnologica Federal do Parana (UTFPR), Pato Branco, Parana, Brazil
[2] Graduate Program of Applied Sciences to Health Products, Universidade Federal Fluminense (UFF), Niteroi, Rio de Janeiro, Brazil
[3] Primary Health Care, Pato Branco Prefecture, Parana, Brazil
[4] Innovation Office, Mass General Brigham Hospital, Cambridge, Massachusetts, United States of America

**E-mail:** erickrodrigues@utfpr.edu.br

**Keywords:** vessel, segmentation, learnings, x-ray, cardiac, angiographic

**Abstract**

This work proposes a pixel-classification approach for vessel segmentation in x-ray angiograms. The proposal uses textural features such as anisotropic diffusion, features based on the Hessian matrix, mathematical morphology and statistics. These features are extracted from the neighborhood of each pixel. The approach also uses the ELEMENT methodology, which consists of creating a pixel-classification controlled by region-growing where the result of the classification affects further classifications of pixels. The Random Forests classifier is used to predict whether the pixel belongs to the vessel structure. The approach achieved the best accuracy in the literature (95.48%) outperforming unsupervised state-of-the-art approaches.

## 1. Introduction

Proper vessel visualization can be a challenging problem. The challenge becomes even worse in the presence of stenoses, i.e., blockage or narrowing of vessels due to conditions such as neoplasm. Computational methods are able to provide enhanced visualizations of these vessels, which not only assist physicians in clinical practice but also enable their automatic segmentation.

Some works in the literature mention that calcium score correlates with angiographic findings [1]. The score can be used to predict diseases or even heart failure when patients undergo angiography. Furthermore, the score is also correlated to the number of affected arteries, which in turn can be analyzed using a vessel enhancement tool.

Automated segmentations can be used to measure the rate of dilation of retinal vessels. This rate is associated to diabetes and other health conditions [2]. Furthermore, these segmentations can also be used in image registration [3], as well as to track, find and evaluate the degree or risk of stenosis.

Stergiopulos *et al* [4] model blood flow in order to automatically locate and quantify the degree or severity of potential stenoses. However, prior to this type of modeling, vessels must be manually or automatically segmented, which is a laborious and repetitive task.

Filters that enhance what is called ‘vesselness’, coupled with heuristics that threshold the filter response, are quite popular unsupervised segmentation approaches. Although some methods such as the very popular Frangi filter proposed in [5] enhance the visualization of different vessels in a broad extent of modalities such as x-rays and Magnetic Resonance Angiograms (MRA), most methods focus on specific modalities and on a single type of vessel (e.g., cardiac vessels, retinal vessels, etc) [6, 7, 8–10, 11] in order to achieve more accurate enhancements or segmentations.

This work focuses on experimenting with the novel ELEMENT framework [12], which combines classical texture and edge feature extraction with a new concept for vessel segmentation called connectivity. The ELEMENT proposal [12] focus on different modalities for retinal vessel images. We propose a different feature set that works appropriately for x-ray angiographic imaging and compare the obtained results with the literature. Vessels in x-ray were not previously evaluated as a potential fit for the ELEMENT methodology.

The ELEMENT methodology [12] works as follows: the classification/estimation of a certain pixel into vessel or non vessel is stored along with the segmentation procedure and influences on the classification of surrounding pixels. This step is crucial and is performed in real-time during classification. A couple of features that represent this connectivity aspect are included in the feature vector and the classification of the pixels are reproduced respecting a region growing approach. As shown in a previous work [12], the outcome is a very robust vessel segmentation.

When it comes to x-ray Angiograms, previous work [6, 8, 13–16] reported results using private datasets lacking a proper comparison and reproducibility of the approaches. In contrast, we create a novel public dataset and provide a ground truth annotated by one of the authors of this work (a physician).

This manuscript is organized as follows: in the next section we perform a literature review, addressing works that propose vessel segmentation or methodologies for vessel enhancement. In section 3, we describe our proposed methodology in detail. Section 4 highlights the obtained results and comparisons. Finally, last section summarizes the work, discusses obtained results and propose avenues for future work.

## 2. Literature review

Vessel segmentation is a problem tackled from many perspectives. Some works provide methods to segment vessels from an intercoronary perspective [17]. Others focus on the 3D segmentation of vessels [18, 19]. This work, in contrast, focus on the 2D segmentation of x-ray angiograms.

Segmentation solutions can either be supervised or unsupervised [20, 21]. When it comes to X-Ray imaging, the works in the current literature use unsupervised methodologies, as opposed to a supervised approach such as the one proposed in this work. Usually, supervised methodologies are capable of obtaining more accurate segmentations [12]. Unlike retinal fundus images, the two key challenges in x-ray angiograms are the low contrast between coronary arteries and image background as well as non-uniform illumination.

Fatemi *et al* [6] apply vesselness filters built upon the Hessian matrix to the entire sequence of x-ray cardiac angiograms. Next, the angiographic frames and their filtered outputs are aggregated using wavelet transforms. The merged image, along with a set of constant coefficients, is applied to the filtered image to obtain the final segmented version.

Mhiri *et al* [8] also combine a Hessian-based vesselness filter to a heuristic, which is called random walk formulation. The heuristic consists of using manually selected seed points to improve and refine segmentation results. Similarly to Fatemi *et al*, the authors also work with x-ray cardiac angiograms. They claim that their methodology achieved better results in comparison to the widely popular Frangi filter.

Sanchez *et al* [14] combine Boltzmann Univariate Marginal Distribution Algorithm (BUMDA) and Gabor filters to produce segmented vessels in x-ray images. Their methodology focuses on optimizing the parameters of the Gabor filter. For this optimization problem, authors use the area under the ROC Curve (AUC) as fitness function. Later, they estimate if a pixel belongs to a vessel branch by using an interclass variance thresholding. The authors claim that their method outperformed all approaches in their comparison.

It is important to highlight that most works that propose vessel segmentation methodologies for x-ray angiograms lack reproducibility. Either they do not provide the dataset, the source code, or both cases, which is the most frequent scenario. Due to that problem, we will briefly shift the attention towards a small review of vessel segmentation in retinal imaging.

The work of Odstrcilik *et al* [9] focuses on retinal imaging. First, brightness and contrast are corrected. Next, the blue and red channels of the images are discarded, which is a common practice in retinal imaging approaches. Matched filtering is then applied to the pre-processed images. Later, convolutions of these filters are applied to the image, considering different rotation angles and orientations. The results are fused to highlight the local maximal responses for each pixel. This highlighted fused version is thresholded to produce a binary map of the vessels. Finally, the map or segmentation is softened using morphological operators [22] to remove small artifacts. Their approach is compared to a total of 12 works, where their methodology was at least as good as the remaining.

Lupascu *et al* [23] also use a pixel-classification approach for retinal fundus images. The authors extract a total of 41 features also from the green layer of the images (red and blue are discarded). Their feature vector is composed by (1) one feature that is based on Staal ridges, (2) six features that indicate vessel likelihood, (3) twenty two features computed from region and boundary information from the Gaussian scale space, (4) two features based on the Frangi filter, (5) eight features sensitive to elongated structures based on second-order detectors, (6) three features based on Lindeberg ridges, (7) one feature based on the Wavelet transform and (8) one feature based on vesselness likelihood using intensity values. A meta Adaboost classifier was used and authors were able to obtain competitive results.

In contrast to [23], the proposal of this work uses a very different set of features. Besides, the modality and type of image is totally different: x-ray angiograms. In addition, we use the novel ELEMENT connectivity approach [12], which is crucial to achieve accurate results. This fact is evident as the connectivity features are usually ranked first in terms of feature weight during classification [12]. Furthermore, pixels must be classified using the information of surrounding previously classified pixels. This mimicks a brush motion and inserts this humanized perspective into the machine learning domain. The pipeline is shown in figure 1.

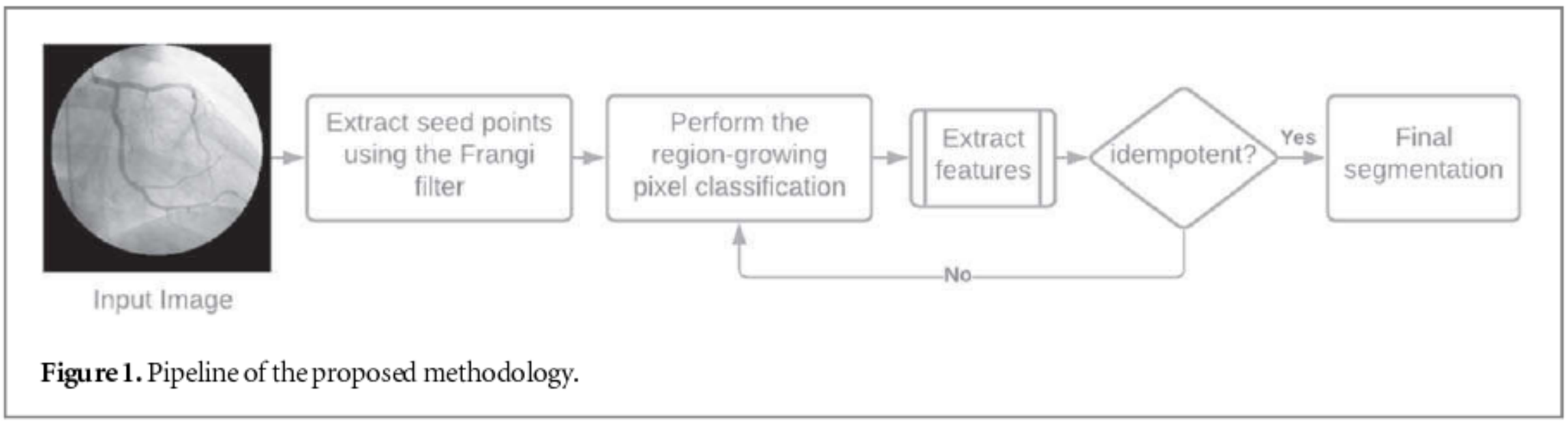

**Figure 1.** Pipeline of the proposed methodology.

## 3. Proposed methodology

The dataset used in this work is provided by OSIRIX and their image library [24]. However, as OSIRIX do not provide a ground truth containing the segmentation, a physician was designed to perform the manual annotations of the ground truth. This ground truth is available at [25]. Figure 2 shows all the original x-ray images (total of 7 images). Figure 3, on the other hand, shows the ground truth images.

Although we do work with a limited set of images (7 at total), this approach is commonplace in the literature for this type of cardiac modality. As our proposal works with pixel-level verifications, we consider a total of 1,835,008 instances ($7 \times 512 \times 512$), which is more than enough for a fair statistical analysis. In this sense, a single image contains a total of 262,144 instances/pixels.

The proposed methodology consists of two steps: (1) feature extraction and (2) region-growing classification. At least one image, along with its ground truth, is required to properly train the classifier. Once the algorithm is trained, the same features are extracted from unprocessed images. The classification model is able to determine whether the pixel under consideration belongs to a vessel branch based on the extracted features.

Figure 1 shows the pipeline of the proposed approach. The connectivity score must be updated in real-time every time a new pixel is classified. The brush motion that humans perform while manually annotating vessels is reproduced by the ELEMENT methodology. This idea is used to enhance the classification accuracy. However, it requires a controlled pixel-classification process, where the information of previously classified vessels is used for the classification of new vessels. If necessary, more specific details can be found in [12].

### 3.1. Feature vector

Each pixel at position $(x, y)$ is described by a feature vector. The proposed feature vector includes features that detect and process grey-level information from the input image as well as a simple yet robust group of features. In what follows, we provide the details for each class of grey-level features. The features consist of indices computed from the Hessian matrix [26], a Laplacian filter [27], morphological opening and closing [3, 21, 28], anisotropic diffusion [29], mean blur, Kuwahara filter [30] and an adaptation of the Sobel filter [31].

*Hessian Matrix (10 Features)* Features built upon the Hessian matrix are very popular in vessel segmentation approaches. The widely popular Frangi filter also falls in this category. A total of 10 Hessian-based features are extracted, where the Hessian matrix $H$ is given by equation (1):

$$H = \begin{bmatrix} I_{xx} & I_{xy} \\ I_{yx} & I_{yy} \end{bmatrix} \tag{1}$$

where $I_{xx}$ and $I_{yy}$ are second-order partial derivatives with respect to $x$ and $y$, respectively. $I_{xy}$ and $I_{yx}$ are second order derivatives with respect to $x$ and $y$ and $y$ and $x$, respectively. The first feature extracted from the Hessian matrix is the determinant, given by equation (2):

$$|H| = ad - cb \tag{2}$$

where $a = I_{xx}$, $b = I_{xy}$, $c = I_{yx}$ and $d = I_{yy}$.

The other 9 features stem from these four elements. The features are, respectively, elements a, b, c and d alone (4 features), the first and second eigenvalues (in equation (3) and 4) the square of the gamma-normalized square eigenvalue difference ($t = 1$) in equation (5), the modulus (Equation (6)) and the trace (equation (7)):

$$\lambda_1 = \frac{-\sqrt[2]{a^2 - 2ad + 4bc + d^2} + a + d}{2} \tag{3}$$

$$\lambda_2 = \frac{\sqrt[2]{a^2 - 2ad + 4bc + d^2} + a + d}{2} \tag{4}$$

$$N_{y-norm}(H) = t^4(a - d)^2((a - d)^2 + 4b^2) \tag{5}$$

$$M(H) = \sqrt[2]{a^2 + bc + d^2} \tag{6}$$

$$tr(H) = a + d \tag{7}$$

*Simple Surrounding Information (4 Features)* For this category we extract the mean, max, min and median values from windows of $7 \times 7$ pixels. This value was chosen empirically and can be increased if the resolution of the images is increased. The value was chosen in order to encompass a substantial part of the vessel caliber. If this value is too great, the performance could be impacted.

*Anisotropic Diffusion (4 Features)* Anisotropic diffusion operations aim to remove or reduce the noise (similar to a blurring effect) while preserving borders, which is something that the usual blur ignores. The anisotropic diffusion is computed with respect to pixel $I(x, y)$, as follows:

$$an(x, y) = I(x, y) + \lambda(\partial(I(x, y-1))I(x, y-1) + \partial(I(x, y+1))I(x, y+1) + \partial(I(x-1, y))I(x-1, y) + \partial(I(x+1, y))I(x+1, y) \quad (8)$$

where the function *an* is repeated for *i* iterations (set according to preference and goal) and $\partial$ represents the flux derivative while $\kappa$ is a constant that controls the sensitivity to edges and is usually chosen experimentally, both shown in equation (9):

$$\partial(x, y) = e^{\left(\frac{I(x,y)}{\kappa}\right)^2} \quad (9)$$

A total of 4 features are extracted based on the anisotropic diffuction, the set of parameters is: $\{(\lambda = 0.3, i = 20, \kappa = 4), (i = 10, \lambda = 0.5, \kappa = 3), (i = 35, \lambda = 2, \kappa = 3), (\kappa = 6, i = 40, \lambda = 0.8)\}$.

*Closing and Opening Operations from Mathematical Morphology (6 Features)* Opening and closing are operations built upon morphological dilations and erosions. An opening is obtained after applying dilations followed by erosions, while closing is the 'opposite' (dual operation), erosions followed by dilations [20, 21].

Dilation and erosion are represented using set theory. In the binary case, the image is defined as a set that contains all the foreground pixels (either black or white pixels, according to preference). Binary dilations are defined as the union of the translation of all elements in image *I* by pixels of the structuring element *B* (which is also an image), as shown in equation (10):

$$dil^B(I) = \bigcup_{b \in B} I_b \quad (10)$$

where $I_b$ represents the image $I$ translated by $b$. That is, suppose that $I = \{(0, 0), (1, 2), (3, 5)\}$ and $b = (1, 0)$, then $I_b = \{(1, 0), (2, 2), (4, 5)\}$.

Erosion is the dilation dual operation. In this case, the union is replaced with the intersection, as shown in equation (12)

$$er^B(I) = \bigcap_{b \in B} I_{-b} \quad (11)$$

As the images in our application are grey level images, we provide the definitions of grey level dilations and erosions. The grey-level dilation of a pixel $I(x, y)$ is given by:

$$dil^B(x, y) = \max_{b \in B}\{I((x, y)_b) + B(b)\} \quad (12)$$

where $(x, y)_b$ represents pixel $(x, y)$ translated by $b$. That is, if $b = (-1, 1)$ and $(x = 8, y = 9)$, then $(x, y)_b$ is equal to the pixel at position $(x = 7, y = 10)$. $I((x, y)_b)$ returns the grey value of this pixel and $B(b)$ returns the grey value of the pixel $b$ of the structuring element.

Similarly, in erosions, the max operator is replaced with the min operator, as shown in equation (13):

$$er^B(x, y) = min_{b \in B}\{I((x, y)_{-b}) + B(-b)\} \quad (13)$$

Finally, the opening and closing operations are defined by equation (14) and 15, respectively:

$$I \circ B = er^B(dil^B(x, y)) \quad (14)$$

$$I \bullet B = dil^B(er^B(x, y)) \quad (15)$$

Two different types of structuring elements were used: $B_1 = \{(0, 255, 0), (255, 255, 255), (0, 255, 0)\}$ and $B_2$ is a 10x10 image containing a central symmetrical circle filled with the value 255 (in a 8-bit image range). A total of 6 features were extracted, defined by the following sets of parameters: $\{(B = B_1, j = 1, i = 1), (B = B_1, j = 1, i = 3), (B = B_1, j = 3, i = 1), (B = B_2, j = 1, i = 1), (B = B_2, j = 1, i = 3), (B = B_2, j = 3, i = 1)\}$, where *i* is the amount of applied erosions and *j* the amount of dilations.

*Kuwahara Filter (2 Features)* Similar to the anisotropic diffusion, the Kuwahara filter removes noise from images while aiming to preserve edges. Let us assume a square window of size $2a + 1$ around the processed pixel, the four quadrants $Q_{i=1...4}$ around this pixel $(x, y)$ are defined as:

$$Q_i(x, y) = \begin{cases} [x, x+a] \times [y, y+a], & \text{if i} = 1 \\ [x-a, x] \times [y, y+a], & \text{if i} = 2 \\ [x-a, x] \times [y-a, y], & \text{if i} = 3 \\ [x, x+a] \times [y-a, y], & \text{if i} = 4 \end{cases} \quad (16)$$

where $\times$ is the cartesian product.

The standard deviation $\sigma_i(x, y)$ and arithmetic mean $m_i(x, y)$ of the four quadrants are used to determine the value of the central pixel. The final output of the Kuwahara filter $\phi(x, y)$ is given by:

$$\phi(x, y) = \begin{cases} m_1(x, y), & \text{if } \sigma_1(x, y) = \min_i\{\sigma_i(x, y)\} \\ m_2(x, y), & \text{if } \sigma_2(x, y) = \min_i\{\sigma_i(x, y)\} \\ m_3(x, y), & \text{if } \sigma_3(x, y) = \min_i\{\sigma_i(x, y)\} \\ m_4(x, y), & \text{if } \sigma_4(x, y) = \min_i\{\sigma_i(x, y)\} \end{cases} \quad (17)$$

The value of the central pixel is the mean value of the area that is most homogeneous. We consider two window sizes for the Kuwahara filter computations, $10 \times 10$ and $20 \times 20$, which correspond to 2 features.

*Rodrigues Sobel Filter (2 Features)* Rodrigues *et al* [31] proposed a slight modification of the Sobel filter that considers distances and shadow/light aspects of the image. We also extract two features based on the light separation.

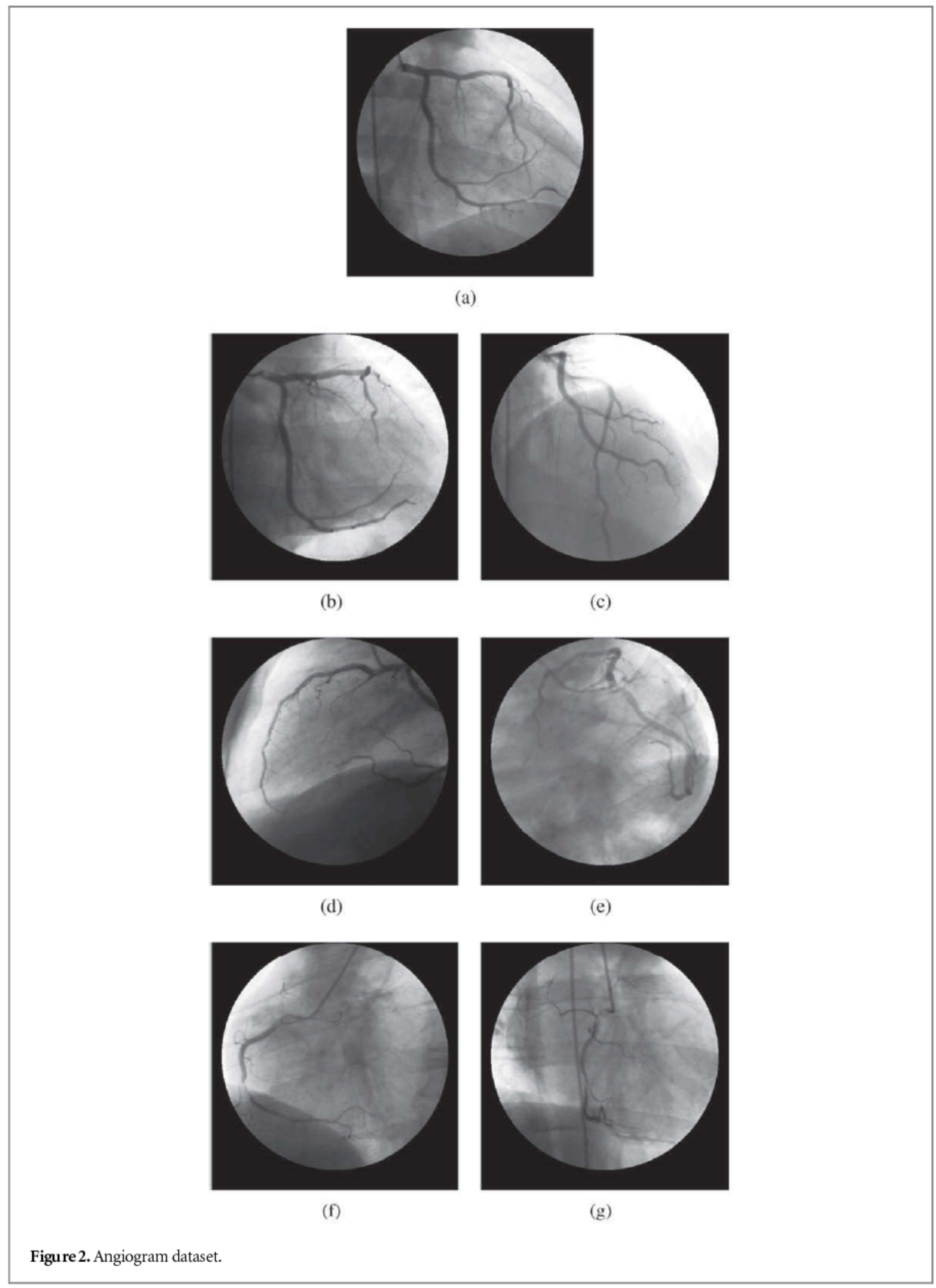

**Figure 2.** Angiogram dataset.

The Light Sobel filter is defined in equation (18).

$$LS(d, t, x, y) = \begin{cases} 1, & \text{if } (l_1 \text{ and } l_2) \\ 0, & \text{otherwise} \end{cases} \tag{18}$$

where

$$\begin{aligned} l_1 &= I(x, y) - I(x, y - d) > t \\ &\text{and } \ I(x, y) - I(x, y + d) > t \\ l_2 &= I(x, y) - I(x - d, y) > t \\ &\text{and } I(x, y) - I(x + d, y) > t \end{aligned} \tag{19}$$

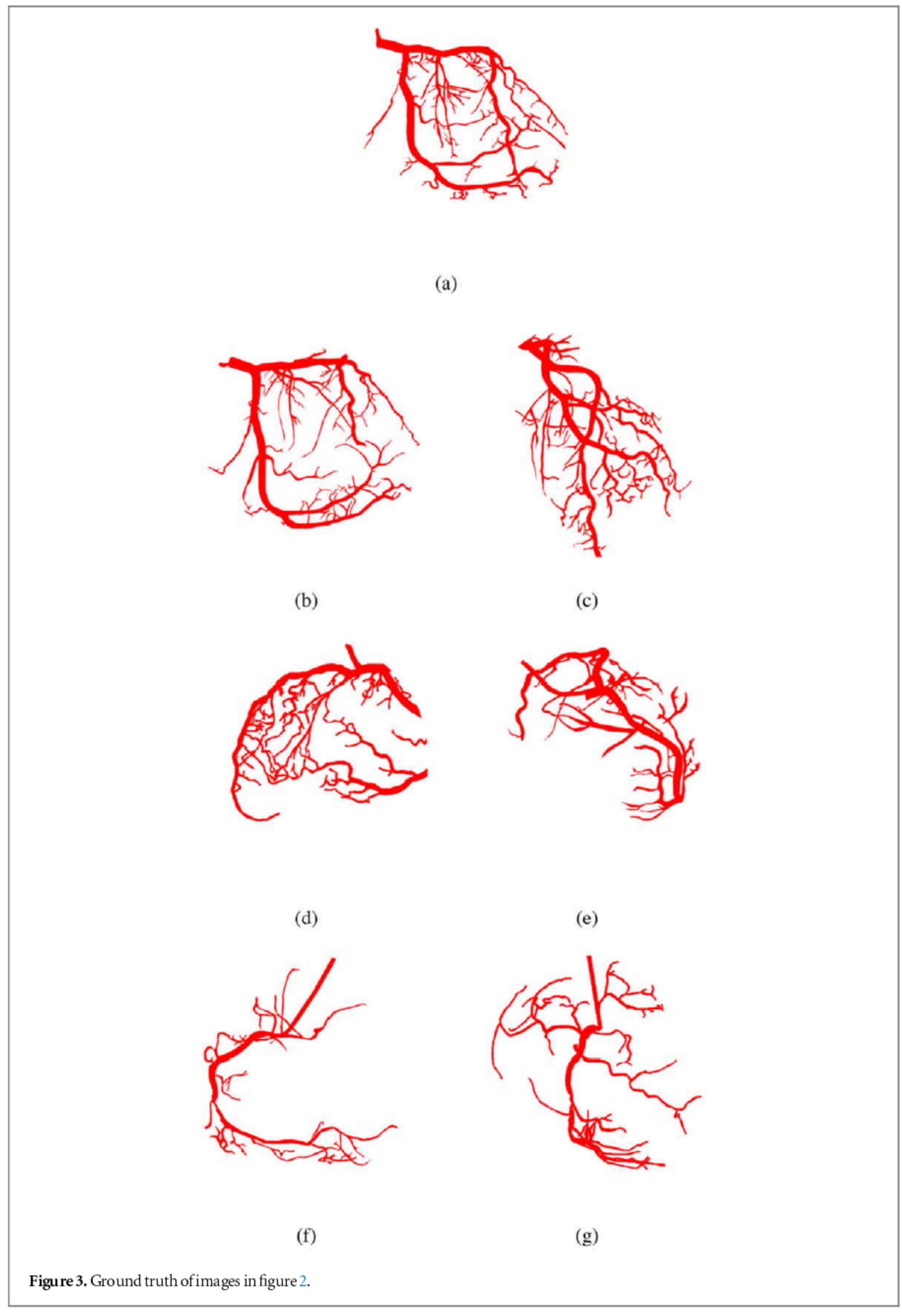

**Figure 3.** Ground truth of images in figure 2.

Parameters $t$ and $d$ were set to these two configurations: ($t = -10$, $d = 2$) and ($t = -10$, $d = 5$).

### 3.2. Connectivity features

Besides the described feature set, we also extract two real-time connectivity features: (1) the immediate connectivity and (2) the radial connectivity feature with radius equal to 7. These two features are flag-like features that indicate if a pixel has been classified as vessel in the neighbourhood.

The idea is that if this assumption is true, it is more likely that the pixel being processed is a vessel pixel as well. These features represent the brush motion. Classifiers will associate this information to the grey-level extracted information and decide whether the pixel belongs to a vessel structure.

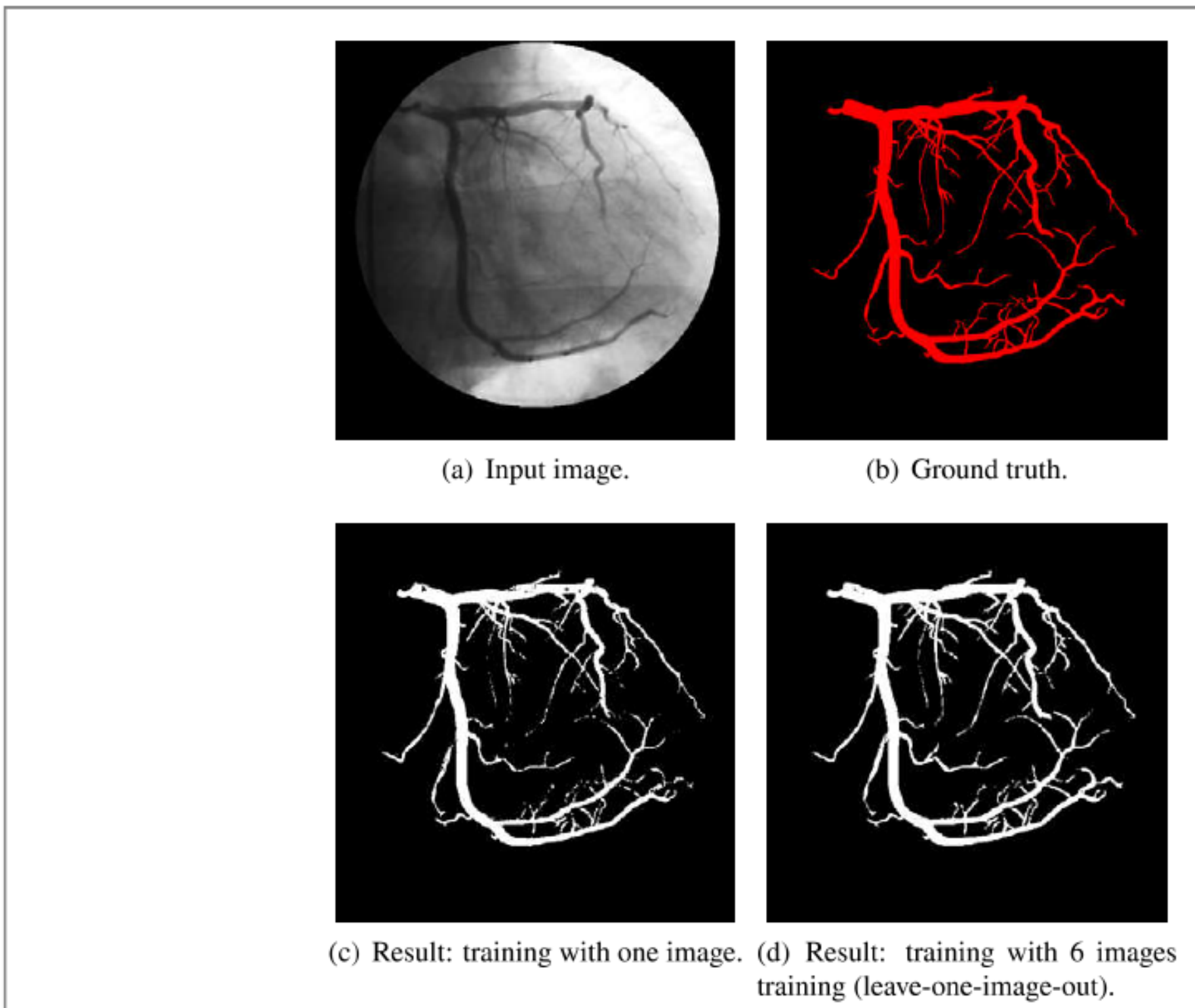

**Figure 4.** X-ray angiogram image with ground truth data and associated segmentation results using the Random Forests classifier.

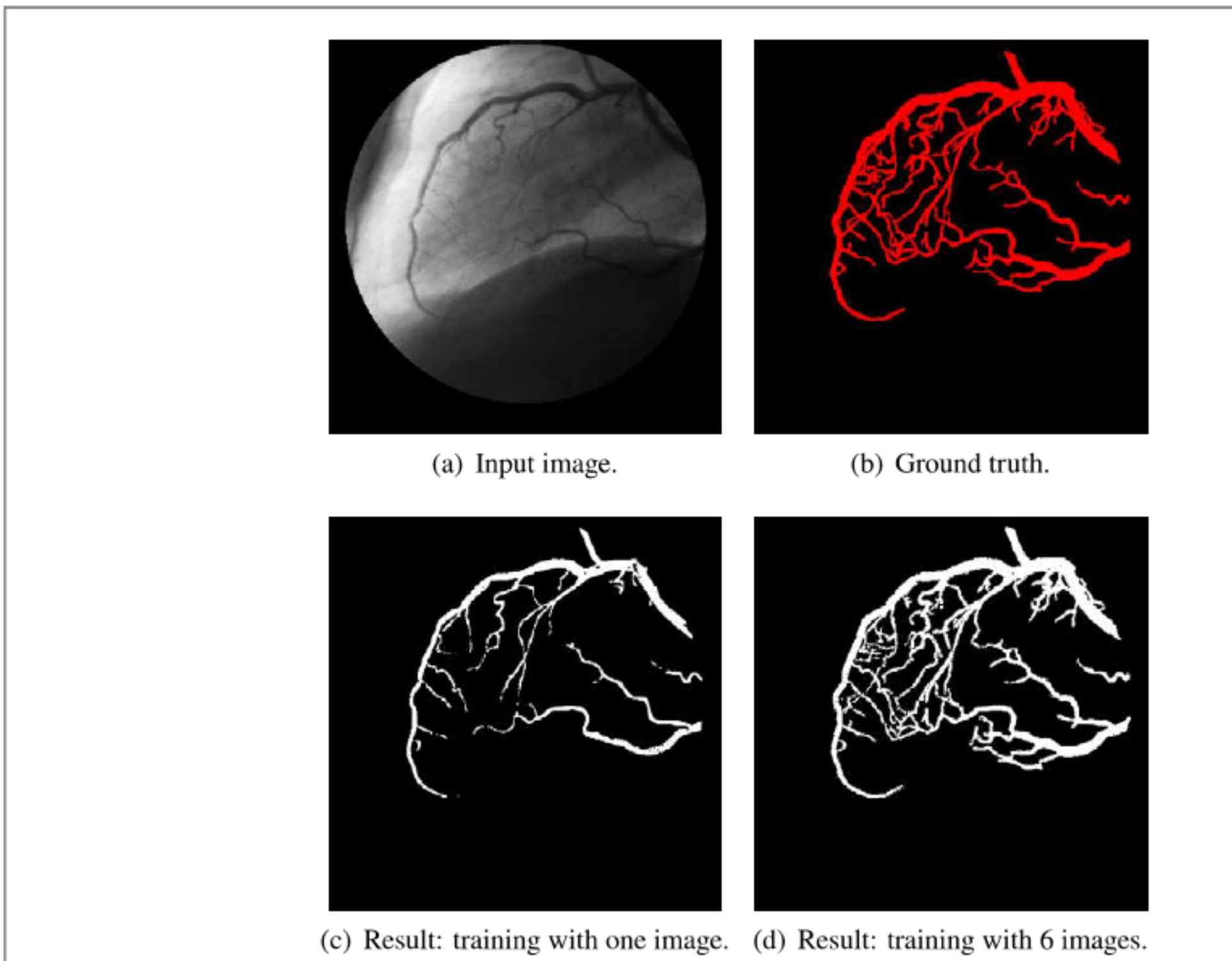

**Figure 5.** X-ray angiogram image with ground truth data and associated segmentation results using the Random Forests classifier.

## 4. Experiments

Several classifiers [32] such as Random Forests, RBF Classifier, SPegasos, REPTree, Naive Bayes, Bayes Net,

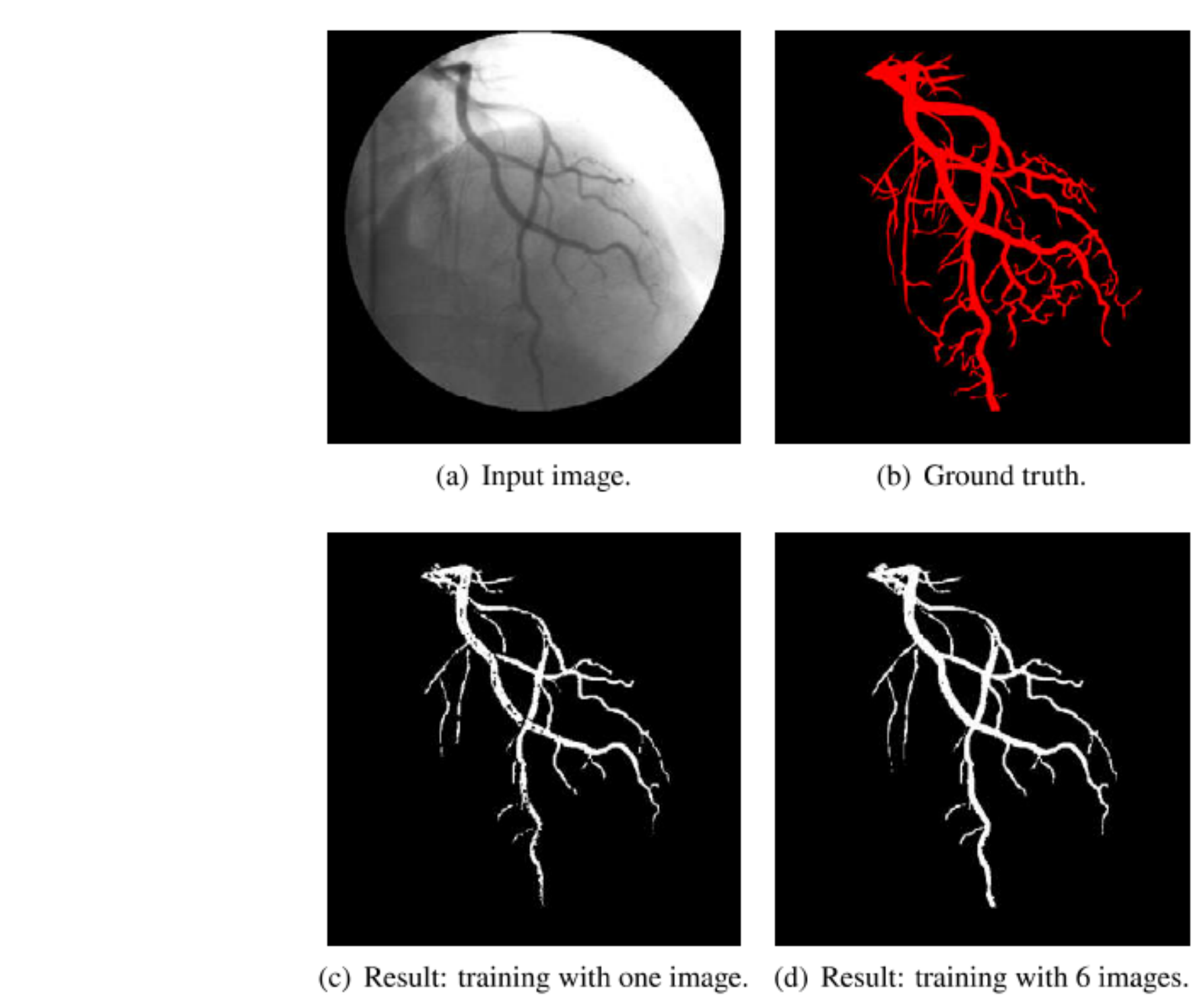

(a) Input image. (b) Ground truth.

(c) Result: training with one image. (d) Result: training with 6 images.

**Figure 6.** X-ray angiogram image with ground truth data and associated segmentation results using the Random Forests classifier.

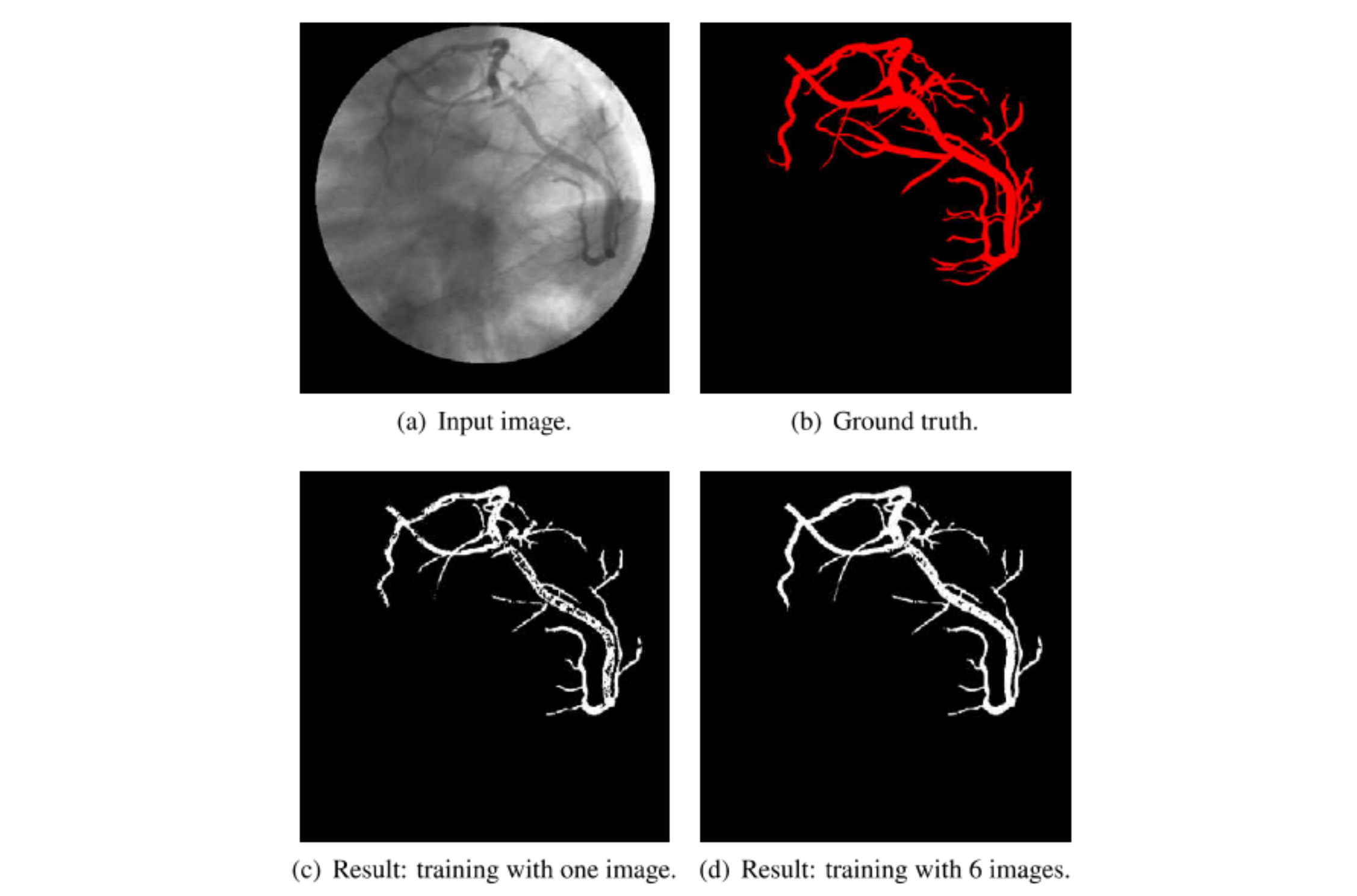

(a) Input image. (b) Ground truth.

(c) Result: training with one image. (d) Result: training with 6 images.

**Figure 7.** X-ray angiogram image with ground truth data and associated segmentation results using the Random Forests classifier.

J48 Graft, Decision table, k-NN [33], Hoeffding Tree from the Weka framework [34] were evaluated. Among these, Random Forests [35] was the classifier that returned the best results and was selected to perform the pixel classification.

Figures 4, 5, 6, and 7 show unprocessed images along with their manual segmentations and automatic segmentations produced by the proposed vessel classification framework. The results shown in figures (c) correspond to a segmentation with a model that was

**Table 1.** Comparison of the results.

| Work | TP(%) | TN(%) | Acc(%) | AUC |
|---|---|---|---|---|
| Mihi *et al* [8] | 66 | ··· | ··· | 0.97 |
| Sanchez *et al* [14] | ··· | ··· | 94.4 | ··· |
| Fatemi *et al* [6] | ··· | ··· | 40.1 | ··· |
| Eiho *et al* [36] apud [14] | ··· | ··· | 91.7 | 0.933 |
| Wang *et al* [37] apud [14] | ··· | ··· | 93.1 | 0.937 |
| Kang *et al* [38] apud [14] | ··· | ··· | 90.5 | 0.916 |
| Chanwimaluang *et al* [39] apud [14] | ··· | ··· | 85.2 | ··· |
| Chaudhuri *et al* 40 apud 14 | ··· | ··· | ··· | 0.9176 |
| This work (leave one image out) | 73.9 | 97.3 | 95.48 | 0.965 |

[a] Accuracy is defined as: (TP + TN)/total instances.
[b] TP is true positive rate and TN is true negative rate.
[c] AUC is the area under the ROC curve.

trained with a single image from the ground truth ($512 \times 512$ pixels).

The images in (d), on the other hand, contain the segmentation results obtained after training the Random Forests classifier using the leave one (image) out approach. That is, all but one image (the one that is being segmented) are used for training. Although the differences between images (c) and (d) are not immediately evident, the numerical results improve when leave one out is used.

Table 1 compares the results to the state-of-the-art. This comparison is not entirely fair because [8, 14] and [6] do not provide their dataset and therefore it is impossible to strictly replicate their results using our dataset. The rest of the works do not even provide the indices, they just propose the unsupervised methodology for vessel segmentation, without evaluating the pixel-wise accuracy. Future works can reference and use our ground truth for a fair comparison.

The results reported in Table 1 were obtained using the Random Forests classifier and the ELEMENT paradigm (region-growing pixel-wise classification). We achieved the best accuracy reported in the literature (95.48%) and also the best true positive rate (73.9%) and AUC (0.965).

## 5. Conclusion and discussion

Vessel segmentation is a widely explored topic in the scientific literature. However, this is still a great challenge and still attracts a lot of effort from scientists. Approaches are currently not very accurate and are specific to modalities and types of vessels. Furthermore, not much effort has been directed towards x-ray cardiac angiograms, in contrast to retinal fundus images, for instance.

This work applies for the first time the ELEMENT methodology to x-ray cardiac angiographic images. Obtained results are very close to the original manual segmentation performed by specialists. The proposal achieved the best accuracy in the literature, 95.48%. Furthermore, it also obtained the best area under the roc curve (AUC): 0.965.

Previous experiments also showed that the connectivity features, which are tightly related to the previously described motion pattern, end up being more important than features based on texture and edge information. Although these features ultimately assist with the segmentation by identifying eventual borders and borderline pixels, what more importantly improves the accuracy is the connectivity features.

We expect this approach to be replicated in different vessel modalities and further improved to obtain even more accurate results. We suspect that the results can be improved with a finer selection and wider consideration of features. As a final remark, the created x-ray angiogram ground truth and the source code for the features described in this work are available at [25]. We also provide the extracted feature vectors for all images of both datasets.

## 6. Disclosure

The authors declare no competing financial interests or personal relationships that could have appeared to influence the work reported in this paper.

## Acknowledgments

L O Rodrigues wants to acknowledge the financial support from CAPES Higher Education Improvement Coordination (financial code 001) in the form of scholarship.

## Data availability statement

The data that support the findings of this study are openly available at the following URL/DOI: https://doi.org/https://github.com/Oyatsumi/X-RayAngiogramDataset.

## ORCID iDs

E O Rodrigues https://orcid.org/0000-0001-9124-3603

## References

[1] Detrano R *et al* 1996 Prognostic value of coronary calcification and angiographic stenoses in patients undergoing coronary angiography *J. Am. College Cardiol.* **27** 285–90

[2] Lim L S, Ling L H, Ong P G, Foulds W, Tai E S and Wong T Y 2017 Dynamic responses in retinal vessel caliber with flicker light stimulation and risk of diabetic retinopathy and its progression *Investigative Ophthalmol. Vis. Sci.* **58** 2449–55

[3] Rodrigues E O, Morais F F C, Morais N A O S, Conci L S, Neto L V and Conci A 2016a A novel approach for the automated segmentation and volume quantification of cardiac fats on computed tomography *Comput. Methods Programs Biomed.* **123** 109–28
[4] Stergiopulos N, Young D F and Rogge T R 1992 Computer simulation of arterial flow with applications to arterial and aortic stenoses *J. Biomech.* **25** 1477–88
[5] Frangi A F, Niessen W J, Vincken K L and Viergever M A 1998 Multiscale vessel enhancement filtering *Medical Image Computing and Computer-Assisted Intervention* **1496** 130–7
[6] Fatemi M J R, Mirhassani S M and Yousefi B 2010 Vessel segmentation in x-ray angiographic images using hessian based vesselness filter and wavelet based image fusion *Proceedings of the 10th IEEE International Conference on Information Technology and Applications in Biomedicine*
[7] Hannik J, Duits R and Bekkers E 2014 Crossing-preserving multi-scale vesselness *Medical Image Computing and Computer-Assisted Intervention* **8674** 603–10
[8] Mhiri F, Duong L, Desrosiers C and Cheriet M 2013 Vesselwalker: coronary arteries segmentation using random walks and hessian-based vesselness filter *Biomedical Imaging (ISBI)* (https://ieeexplore.ieee.org/document/5687605)
[9] Odstrcilik J, Kolar R, Budai A, Hornegger J, Jan J, Gazarek J, Kubena T, Cernosek P, Svoboda O and Angelopoulou E 2013 Retinal vessel segmentation by improved matched filtering: evaluation on a new high-resolution fundus image database *IET Image Proc.* **7** 373–83
[10] Eibe Frank, Mark A. Hall and Ian H. Witten 2016 Evaluation of hessian-based filters to enhance the axis of coronary arteries in ct images *The WEKA Workbench. Online Appendix for 'Data Mining: Practical Machine Learning Tools and Techniques', Morgan Kaufmann,* Fourth Edition, 2016
[11] Salem N M and Nandi A K 2008 Unsupervised segmentation of retinal blood vessels using a single parameter vesselness measure *Computer Vision, Graphics and Image Processing*
[12] Rodrigues E O, Conci A and Liatsis P 2020 Element: Multi-modal retinal vessel segmentation based on a coupled region growing and machine learning approach *IEEE Journal of Biomedical and Health Informatics* **24** 3507–19
[13] Dehkordi M, Hoseini A M, Sadri S and Soltanianzadeh H 2014 Local feature fitting active contour for segmenting vessels in angiograms *IET Comput. Vision* **8** 167–70
[14] Sanchez F C, Aceves I C, Aguirre A H, Cervantes J G, Meza S S, Rodriguez M O and Cisneros M T 2016 Segmentation of coronary angiograms using gabor filters and boltzmann univariate marginal distribution algorithm *Computational Intelligence and Neuroscience 2016*
[15] Stanfield S A 1986 Angy: A rule-based expert system for automatic segmentation of coronary vessels from digital subtracted angiograms *IEEE Trans. Pattern Anal. Mach. Intell.* **8** 188–99
[16] Zheng Y, Loziczonek M, Georgescu B, Zhou S K, Vega-Higuera F and Comaniciu D 2011 Machine learning based vesselness measurement for coronary artery segmentation in cardiac ct volumes *Proc. SPIE 7962, Medical Imaging 2011: Image Processing*
[17] Gao Z, Chung J, Abdelrazek M, Leung S, Hau W K, Xian Z, Zhang H and Li S 2020a Privileged modality distillation for vessel border detection in intracoronary imaging *IEEE Trans Med Imaging* **39** 1524–34
[18] Gao Z, Liu X, Qi S, Wu W, Hau W K and Zhang H 2017 Automatic segmentation of coronary tree in ct angiography images *Int. J. Adapt Control Signal Process.*
[19] Gao Z, Wang X, Sun S, Wu D, Bai J, Yin Y, Liu X, Zhang H and Albuquerque V H C 2020b Learning physical properties in complex visual scenes: An intelligent machine for perceiving blood flow dynamics from static ct angiography imaging *Neural Netw.* **123** 82–93
[20] Rodrigues E O, Conci A and Liatsis P 2018 Morphological classifiers *Pattern Recognit.* **84** 82–96
[21] Rodrigues E O, Torok L, Liatsis P, Viterbo J and Conci A 2017b k-ms: A novel clustering algorithm based on morphological reconstruction *Pattern Recognit.* **66** 392–403
[22] Rodrigues E O 2021 An efficient and locality-oriented hausdorff distance algorithm: Proposal and analysis of paradigms and implementations *Pattern Recognit.* **117** 66-71
[23] Lupascu C A, Tegolo D and Trucco E 2010 Fabc: Retinal vessel segmentation using adaboost *IEEE Trans. Inf. Technol. Biomed.* **14** 1267–74
[24] OSIRIX, 2020. Dicom image library (https://www.osirix-viewer.com/resources/dicom-image-library/)
[25] Rodrigues E O 2020 X-ray angiograms ground truth (https://github.com/Oyatsumi/X-RayAngiogramDataset)
[26] Lorenz C, Carlsen C, Buzug T M, Fassnacht C and Weese J 2005 A multi-scale line filter with automatic scale selection based on the hessian matrix for medical image segmentation *Scale-Space Theory in Computer Vision* **1252** 152–63
[27] Ilk H G, Jane O and Ilk O 2011 The effect of laplacian filter in adaptive unsharp masking for infrared image enhancement *Infrared Phys. Technol.* **54** 427–38
[28] Vincent L 1994 Morphological area openings and closings for grey-scale images *Shape in Picture* **126** 197–208
[29] Perona P and Malik J 1990 Scale-space and edge detection using anisotropic diffusion *IEEE Trans. Pattern Anal. Mach. Intell.* **12** 629–39
[30] Bartyzel K 2016 Adaptive kuwahara filter *Signal, Image and Video Processing* **10** 663–70
[31] Rodrigues E O, Porcino T M, Conci A and Silva A C 2016b A simple approach for biometrics: Finger-knuckle prints recognition based on a sobel filter and similarity measures *International Conference on Systems, Signals and Image Processing (IWSSIP)*
[32] Rodrigues E O, Pinheiro V H A, Liatsis P and Conci A 2017a Machine learning in the prediction of cardiac epicardial and mediastinal fat volumes *Comput. Biol. Med.* **89** 520–9
[33] Rodrigues E O 2018 Combining minkowski and cheyshev: New distance proposal and survey of distance metrics using k-nearest neighbours classifier *Pattern Recognit. Lett.* **110** 66–71
[34] Frank E, Hall M A and Witten I H 2016 The weka workbench, online appendix for data mining: practical machine learning tools and techniques *Morgan Kaufmann*
[35] Breiman L 2001 Random forests *Mach. Learn.* **45** 5–32
[36] Eiho S and Qian Y 1997 Detection of coronary artery tree using morphological operator *Computers in Cardiology*
[37] Wang S, Li B and Zhou S 2012 A segmentation method of coronary angiograms based on multi-scale filtering and region-growing *International Conference on Biomedical Engineering and Biotechnology*
[38] Kang W, Kang W, Li Y and Wang Q 2013 *The segmentation method of degree-based fusion algorithm for coronary angiogramsInternational Conference on Measurement, Information and Control*
[39] Chanwimaluang T, Fan G and Fransen S R 2006 Hybrid retinal image registration *IEEE Trans. Inf. Technol. Biomed.* **10** 129–42
[40] Chaudhuri S, Chatterjee S, Katz N, Nelson M and Goldbaum M 1989 Detection of blood vessels in retinal images using two-dimensional matched filters *IEEE Trans. Med. Imaging* **8** 263–9